# Synthetic vs. Real Reference Strings for Citation Parsing, and the Importance of Re-training and Out-Of-Sample Data for Meaningful Evaluations: Experiments with GROBID, GIANT and Cora

Mark Grennan
Trinity College Dublin, School of Computer Science and Statistics
Ireland
grennama@tcd.ie

Joeran Beel
Trinity College Dublin, School of Computer Science and Statistics,
Artificial Intelligence Discipline,
ADAPT Research Centre
Ireland
beelj@tcd.ie

## ABSTRACT

Citation parsing, particularly with deep neural networks, suffers from a lack of training data as available datasets typically contain only a few thousand training instances. Manually labelling citation strings is very time-consuming, hence synthetically created training data could be a solution. However, as of now, it is unknown if synthetically created reference-strings are suitable to train machine learning algorithms for citation parsing. To find out, we train Grobid, which uses Conditional Random Fields, with a) human-labelled reference strings from 'real' bibliographies and b) synthetically created reference strings from the GIANT dataset. We find that both synthetic and organic reference strings are equally suited for training Grobid (F1 = 0.74). We additionally find that retraining Grobid has a notable impact on its performance, for both synthetic and real data (+30% in F1). Having as many types of labelled fields as possible during training also improves effectiveness, even if these fields are not available in the evaluation data (+13.5% F1). We conclude that synthetic data is suitable for training (deep) citation parsing models. We further suggest that in future evaluations of reference parsers both evaluation data similar and dissimilar to the training data should be used for more meaningful evaluations.

---

[1] The work presented in this manuscript is based on Mark Grennan's Master thesis "1 Billion Citation Dataset and Deep Learning Citation Extraction" at Trinity College Dublin, Ireland, 2018/2019

## 1 INTRODUCTION[1]

Accurate citation data is needed by publishers, academic search engines, citation & research-paper recommender systems and others to calculate impact metrics [3, 21], rank search results [5, 6] generate recommendations [4, 11–13, 22, 25] and other applications e.g. in the field of bibliometric-enhanced information retrieval [8]. Citation data is typically parsed from unstructured non-machine-readable text, which often originates from bibliographies found in PDF files on the Web (Figure 1). To facilitate the parsing process, a dozen [38] open source tools were developed including *ParsCit* [10], *Grobid* [26, 27], and *Cermine* [35], with Grobid typically being considered the most effective one [38]. There is ongoing research that continuously leads to novel citation-parsing algorithms including deep learning algorithms [1, 7, 30–33, 41] and meta-learned ensembles [39, 40].

Most citation parsing tools apply supervised machine learning [38]. Consequently, labelled data is required to train the algorithms. However, training data is rare compared to other disciplines where datasets may have millions of instances. To the best of our knowledge, existing citation-parsing datasets typically contain a few thousand instances and are domain specific (Table 1). This may be sufficient for traditional machine learning algorithms but not for deep learning, which shows a lot of potential for citation parsing [1,

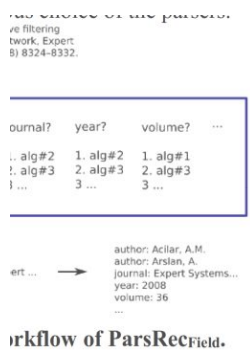

**Figure 1: Illustration of a 'Bibliography' with four 'Reference Strings', each with a number of 'Fields'. A reference parser receives a reference string as input, and outputs labelled fields, e.g. `<author>C. Lemke<\author>` … `<title> Metalearning: a survey` … `<\title>` …**

30–33, 41]. Even for traditional machine learning, existing datasets may not be ideal as they often lack diversity in terms of citation styles.

Recently, we published *GIANT*, a synthetic dataset with nearly 1 billion annotated reference strings [19]. More precisely, the dataset contains 677,000 unique reference strings, each in around 1,500 citation styles (e.g. APA, Harvard, ACM). The dataset was synthetically created. This means, the reference strings are not 'real' reference strings extracted from 'real' bibliographies. Instead, we downloaded 677,000 references in XML format from CrossRef, and used Citeproc-JS [14] with 1,500 citation styles to convert the 677,000 references into a total of 1 billion annotated citation strings (1,500 * 677,000)[2].

We wonder how suitable a synthetic dataset like GIANT is to train machine learning models for citation parsing. Therefore, we pursue the following research question:

1. How will citation parsing perform when trained on synthetic data, compared to being trained on real reference strings?

Potentially, synthetic data could lead to higher parsing performance, as there is more data and more diverse data (more citation styles). Synthetic data like *GIANT* could also revolutionize (deep) citation parsing, which currently suffers from a lack of 'real' annotated bibliographies at large scale.

In addition to the above research question, we aimed to answer the following questions:

2. To what extent does citation-parsing (based on machine learning) depend on the amount of training data?
3. How important is re-training a citation parser for the specific data it should be used on? Or, in other words, how does performance vary if the test data differs (not) from the training data?
4. Is it important to have many different fields (author, year, …) for training, even if the fields are not available in the final data?

## 2 RELATED WORK

We are aware of eleven datasets with annotated reference strings, the most popular ones probably being Cora and CiteSeer, and authors also often use variations of PubMed (Table 1). Several datasets are from the same authors, and many datasets include data from other datasets. For instance, the Grobid dataset is based on some data from Cora, PubMed, and others [28]. New data is continuously added to Grobid's dataset. As such, there is not "*the one*" Grobid dataset. GIANT is the largest and most diverse dataset in terms of citation styles, but GIANT is, as mentioned, synthetically created.

Cora is one of the most widely used datasets but has potential shortcomings [2, 10, 31]. Cora is homogeneous with citation strings only from Computer Science. It is relatively small and only has labels for "coarse-grained fields" [2]. For example, the author field does not label each author separately. Prasad et al. conclude that a "shortcoming of [citation parsing research] is that the evaluations have been largely limited to the Cora dataset, which is [...] unrepresentative of the multilingual, multidisciplinary scholastic reality" [31].

---

[2] We use the terms 'citation parsing', 'reference parsing', and 'reference-string parsing' interchangeably.

Table 1: List of Citation Datasets

| Dataset Name | # Instances | Domain |
| --- | --- | --- |
| Cora [29] | 1,295 | Computer Science |
| CiteSeer [16] | 1,563 | Artificial Intelligence |
| Umass [2] | 1,829 | STEM |
| FLUX-CiM CS [20] | 300 | Computer Science |
| FLUX-CiM HS [20] | 2,000 | Health Science |
| GROBID [26–28] | 6,835 | Multi-Domain (Cora, arXiv, PubMed...) |
| PubMed (Central) [9, 17] | Varies | Biomedical |
| GROTOAP2 (Cermine) [35–37] | 6,858 | Biomedical & Computer Science |
| CS-SW [20] | 578 | Semantic Web Conferences |
| Venice [33] | 40,000 | Humanities |
| GIANT [19] | 991 million | Multi-Domain (~1,500 Citation Styles) |

## 3 METHODOLOGY

To compare the effectiveness of synthetic vs. real bibliographies, we used Grobid. Grobid is the most effective citation parsing tool [38] and, based on our experience, one of the most easy to use ones. Grobid uses conditional random fields (CRF) as a machine learning algorithm. Of course, in the long-run, it would be good to conduct experiments with different machine learning algorithms, particularly deep learning algorithms, but for now we concentrate on one tool and algorithm. Given that all major citation-parsing tools -- including Grobid, Cermine and ParsCit – use CRF we consider this sufficient for an initial experiment. Also, we attempted to re-train *Neural ParsCit* [31] but failed doing so, which indicates that the ease-of-use of the rather new deep-learning methods is not yet as advanced as the established citation parsing tools like Grobid.

We trained Grobid, the CRF respectively, on two datasets. *Train$_{Grobid}$* denotes a model trained on 70% (5,460 instances) of the dataset that Grobid uses to train its out-of-the box version. We slightly modified the dataset, i.e. we removed labels for 'pubPlace', 'note' and 'institution' as this information is not contained in GIANT, and hence a model trained on GIANT could not identify these labels[3]. *Train$_{GIANT}$* denotes the model trained on a random sample (5,460 instances) of GIANT's 991,411,100 labeled reference strings. Our expectation was that both models would perform similar, or, ideally, *Train$_{GIANT}$* would even outperform *Train$_{Grobid}$*.

To analyze how the amount of training data affects performance, we additionally trained Train$_{GIANT}$, on 1k, 3k, 5k, 10k, 20k, and 40k instances of GIANT.

We evaluated all models on four datasets. **Eval$_{Grobid}$** comprises of the remaining 30% of Grobid's dataset (2,340 reference strings). *Eval$_{Cora}$* denotes the Cora dataset, which comprises, after some cleaning, of 1,148 labelled reference strings from the computer science domain. *Eval$_{GIANT}$* comprises of 5,000 random reference strings from GIANT.

These three evaluation datasets are potentially not ideal as evaluations are likely biased towards one of the trained models. Evaluating the models on *Eval$_{GIANT}$* likely favors *Train$_{GIANT}$* since the data for both *Train$_{GIANT}$* and *Eval$_{GIANT}$* is highly similar, i.e. it originates from the same dataset. Similarly, evaluating the models on Eval$_{Grobid}$ likely favors Train$_{Grobid}$ as Train$_{Grobid}$ was trained on 70% of the original Grobid dataset and this 70% of the data is highly similar to the remaining 30% that we used for the evaluation. Also, the Cora dataset is somewhat biased, because Grobid's dataset contains parts of Cora. We therefore created another evaluation dataset.

**Eval$_{WebPDF}$** is our 'unbiased' dataset with 300 manually annotated citation strings from PDFs found on the Web. To create Eval$_{WebPDF}$, we chose twenty different words from the homepages of some universities[4]. Then, we used each of the twenty words as a search term in Google Scholar. From each of these searches, we downloaded the first four available PDFs. Of each PDF, we randomly chose four citation strings. This gave approximately sixteen citation strings for each of the twenty keywords and in total, there were 300 citation strings. We consider this dataset to be a realistic, though relatively small, dataset for citation parsing in the context of a web-based academic search engine or recommender system.

We measure performance of all models with precision, recall, F1 (Micro Average) and F1 (Macro Average) on both field level and token level. We only report 'F1 Macro Average on field level' as all metrics led to similar results.

All source code, data (including the WebPDF dataset), images, and an Excel sheet with all results (including precision and recall and token level results) is available on GitHub https://github.com/BeelGroup/GIANT-The-1-Billion-Annotated-Synthetic-Bibliographic-Reference-String-Dataset/.

## 4 RESULTS

The models trained on Grobid (*Train$_{Grobid}$*) and GIANT (*Train$_{GIANT}$*) perform as expected when evaluated on the three 'biased' datasets *Eval$_{Grobid}$*, *Eval$_{Cora}$* and *Eval$_{GIANT}$* (Figure 2). When evaluated on *Eval$_{Grobid}$*, *Train$_{Grobid}$* outperforms *Train$_{GIANT}$* by 35% with an F1 of 0.93 vs. 0.69. When evaluated on *Eval<sub>GIANT</sub>*, results are almost exactly the opposite: *Train<sub>GIANT</sub>* outperforms *Train<sub>Grobid</sub>* by 32% with an F1 of 0.91 vs. 0.69. On *Eval<sub>Cora</sub>*, the difference is less strong but still notable. Train<sub>Grobid</sub> outperforms Train<sub>GIANT</sub> by 19% with an F1 of 0.74 vs. 0.62. This is not surprising as Grobid's training data includes some Cora data. While these results generally might not be surprising, they imply that both synthetic and real data lead to very similar results and 'behave' similarly when used to train models that are evaluated on data being (not) similar to the training data.

---

[3] This is a shortcoming of GIANT. However, the purpose of our current work is to generally compare 'real' vs synthetic data. Hence, both datasets should be as similar as possible in terms of available fields to make a fair comparison. Therefore, we removed all fields that were not present in both datasets.

[4] The words were: bone, recommender systems, running, war, crop, monetary, migration, imprisonment, hubble, obstetrics, photonics, carbon, cellulose, evolutionary, revolutionary, paleobiology, penal, leadership, soil, musicology.

Also interesting is the evaluation on the WebPDF dataset. The model trained on synthetic data (Train<sub>GIANT</sub>) and the model trained on real data (Train<sub>Grobid</sub>) perform alike with an F1 of 0.74 each (Figure 2)[5]. In other words, synthetic and human-labelled data perform equally well for training our machine learning model.

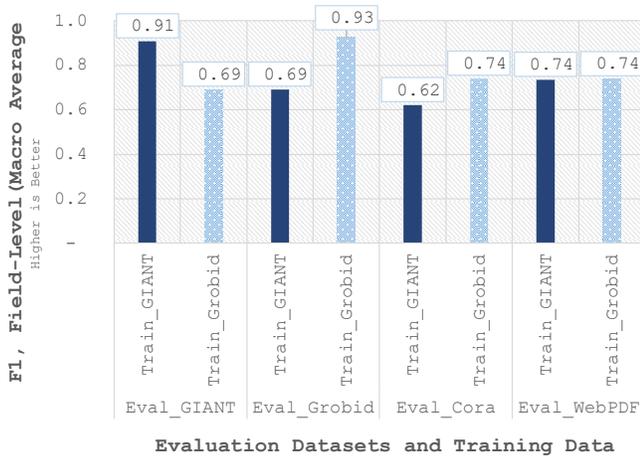

**Figure 2: F1 of the two models (Train<sub>Grobid</sub> and Train<sub>GIANT</sub>) on the four evaluation datasets.**

Looking at the data in more detail reveals that some fields are easier to parse than others (Figure 3). For instance, the 'date' field (i.e. year of publication) has a constantly high F1 across all models and evaluation datasets (min=0.86; max=1.0). The 'author' field also has a high F1 throughout all experiments (min=0.75; max=0.99). In contrast, parsing 'booktitle' and 'publisher' seems to strongly benefit from training based on samples similar to the evaluation dataset. When the evaluation data is similar to the training data (e.g. *Train<sub>GIANT</sub>*--*Eval<sub>GIANT</sub>* or *Train<sub>Grobid</sub>*--*Eval<sub>Grobid</sub>*), F1 is relatively high (typically above 0.7). If the evaluation data is different (e.g. *Train<sub>GIANT</sub>*-- *Eval<sub>Grobid</sub>*), F1 is low (0.15 and 0.16 for Train<sub>Grobid</sub> and Train<sub>GIANT</sub> respectively on Eval<sub>WebPDF</sub>). The difference in F1 for parsing the book-title is around factor 6.5, with an F1 of 0.97 (Train<sub>Grobid</sub>) and 0.15 respectively (*Train<sub>GIANT</sub>*) when evaluated on *Eval<sub>Grobid</sub>*.

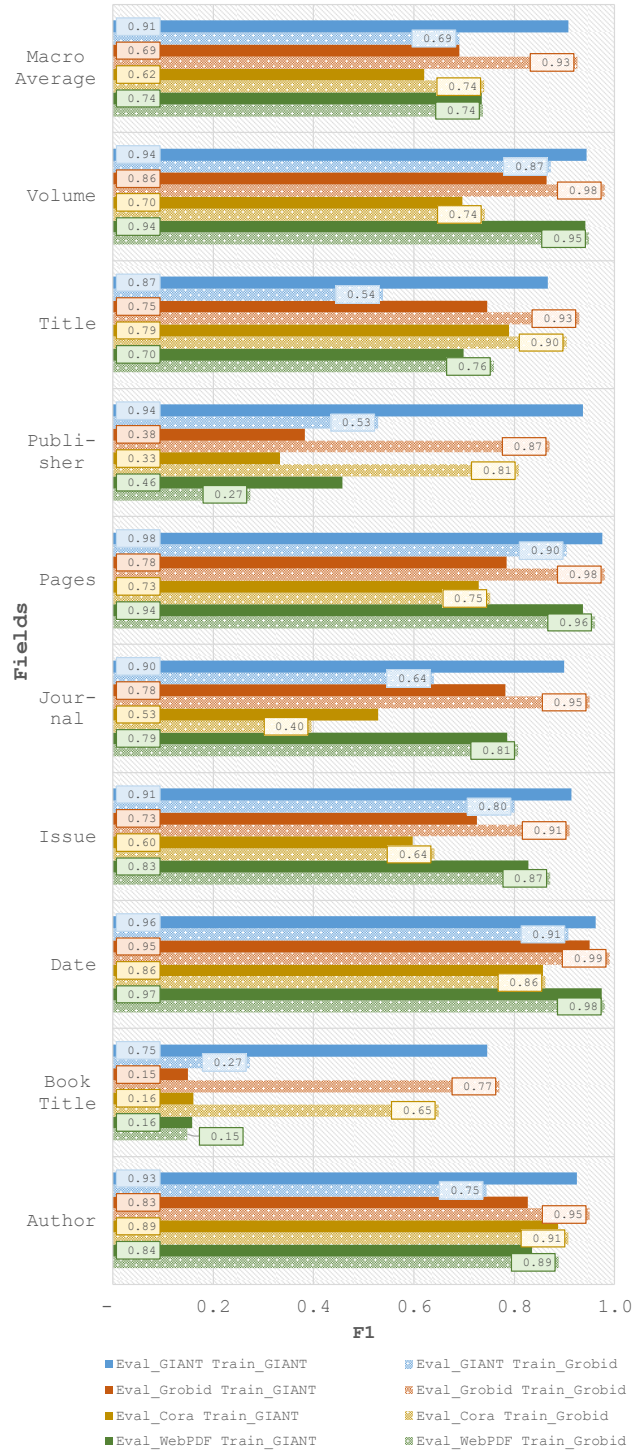

**Figure 3: F1 for different fields (title, author, …), evaluation dataset and training data.**

---

[5] All results are based on the Macro Average F1. Looking at the Micro Average F1 shows a slightly better performance for Train<sub>Grobid</sub> than for Train<sub>GIANT</sub> (0.82 vs. 0.80), but the difference is neither large nor statistically significant (p<0.05).

Similarly, F1 for parsing the book-title on $Eval_{GIANT}$ differs by around factor 3 with an F1 of 0.75 ($Train_{GIANT}$) and 0.27 ($Train_{Grobid}$) respectively. While it is well known, and quite intuitive, that different fields are differently difficult to parse, we are first to show that field accuracy varies for different fields differently depending on whether or not the model was trained on data (not) being similar to the evaluation data.

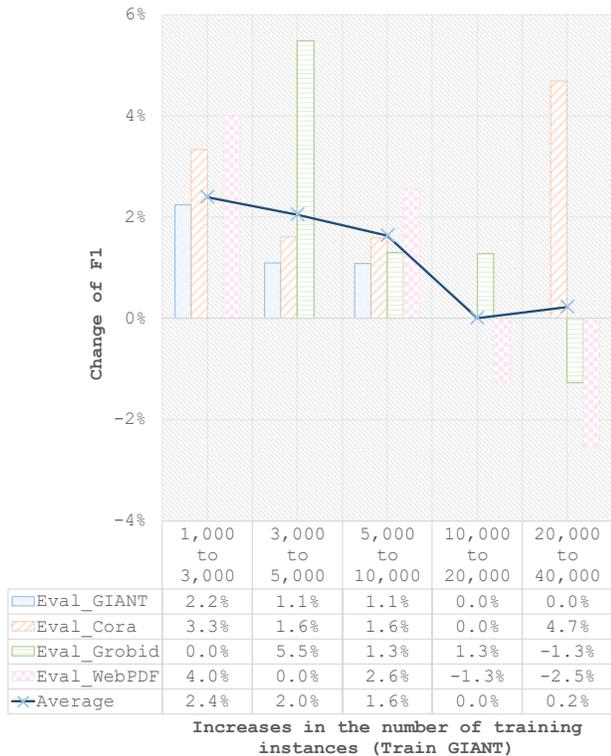

**Figure 4: Performance (F1) of $Train_{GIANT}$ on the four evaluation datasets, by the number of training instances.**

In a side experiment, we trained a new model $Train_{Grobid+}$ with additional labels for institution, note and pubPlace (those we removed for the other experiments). $Train_{Grobid+}$ outperformed $Train_{Grobid}$ notably with an F1 of 0.84 vs. 0.74 (+13.5%) when evaluated on $Eval_{WebPDF}$. This indicates that the more fields are available for training, the better the parsing of all fields becomes even if the additional fields are not in the evaluation data. This finding seems plausible to us and confirms statements by *Anzaroot and McCallum* [2] but, to the best of our knowledge, we are first to quantify the benefit. It is worth noting that citation parsers do not always use the same fields (Table 2). For instance, Cermine extracts relatively few fields, but is one of few tools extracting the *DOI* field.

Our assumption that more training data would generally lead to better parsing performance – and hence GIANT could be useful for training standard machine learning algorithms – was not confirmed. Increasing training data from 1,000 to 10,000 instances improved F1 by 6% on average over all four evaluation datasets (Figure 4). More precisely, increasing data from 1,000 to 3,000 instances improved F1, on average, by 2.4%; Increasing from 3,000 to 5,000 instances improved F1 by another 2%; Increasing further to 10,000 instances improved F1 by another 1.6%. However, increasing to 20,000 or 40,000 instances leads to no notable improvement, and in some cases even to a decline in F1 (Figure 4).

## 5 SUMMARY & DISCUSSION

In summary, both models – the one trained on synthetic data (GIANT) and the one trained on 'real' reference strings annotated by humans (Grobid) – performed very similar. On the main evaluation dataset (WebPDF) both models achieved an F1 of 0.74. Similarly, if a model was trained on data different from its evaluation data, F1 was between 0.6 and 0.7. If a model was trained on data similar to the evaluation data, F1 was above 0.9 (+30%). F1 only increased up to a training size of around 10,000 instances (+6% compared to 1,000 instances). Additional fields (e.g. pubplace) in the training data increased F1 notably (+13.5%), even if these additional fields were not in the evaluation data.

These results lead us to the following conclusions. First, there seems to be little benefit in using synthetic data (i.e. GIANT) for training traditional machine learning models (i.e. conditional random fields). The existing datasets with a few thousand training instances seem sufficient.

**Table 2: The approach and extracted fields of six popular open-source citation parsing tools**

| Citation Parser | Approach | Extracted Fields |
| --- | --- | --- |
| Biblio | Regular Expressions | author, date, editor, genre, issue, pages, publisher, title, volume, year |
| BibPro | Template Matching | author, title, venue, volume, issue, page, date, journal, booktitle, techReport |
| CERMINE | Machine Learning (CRF) | author, issue, pages, title, volume, year, DOI, ISSN |
| GROBID |  | authors, booktitle, date, editor, issue, journal, location, note, pages, publisher, title, volume, web, institution |
| ParsCit |  | author, booktitle, date, editor, institution, journal, location, note, pages, publisher, tech, title, volume |
| Neural ParsCit | Deep Learning | author, booktitle, date, editor, institution, journal, location, note, pages, publisher, tech, title, volume |

Second, citation parsers should, if possible, be (re)trained on data that is similar to the data that should actually be parsed.

Such a re-training increased performance by around 30% in our experiments. This finding may explain why researchers often report excellent performance of their tools and approaches with e.g. F1's of over 0.9. These researchers typically evaluate their models on data highly similar to the training data. This might be considered a realistic scenario for those cases when re-training is possible. However, reporting such results creates unrealistic expectations for scenarios without the option to re-train, i.e. for users who just want to use a citation parser like Grobid out-of-the-box. Therefore, we propose that future evaluations of citation parsing algorithms should be conducted on at least two datasets: One dataset that is similar to the training dataset, and one out-of-sample dataset that differs from the training data.

Third, citation parsers should be trained with as many labelled field types as possible, even if these fields will not be in the data that should be parsed. Such a fine-grained training improved F1 by 13.5% in our experiments.

Fourth, having ten times as much training data (10,000 vs. 1,000) improved the parsing performance by 6%, without notable improvements beyond 10,000 instances. Annotating a few thousand instances should be feasible for many scenarios. Hence, businesses and organizations who want the maximum accuracy should annotate their own data for training as this likely will lead to large increases in accuracy (+30%, see conclusion 3).

Fifth, given how similar synthetic and traditionally annotated data perform, synthetic data likely is suitable to train deep neural networks for citation parsing. This, of course, has yet to be empirically to be shown. However, if our assumption holds true, deep citation parsers could greatly benefit from synthetic data like GIANT.

For the future, we see the need to extend our experiments to different machine learning algorithms and datasets (e.g. unarXive [34] or CORE [23]). It would also be interesting to analyze if and to what extend synthetic data could improve related disciplines. This may include citation-string matching, i.e. analyzing whether two different reference strings refer to the same document [15], or the extraction of mathematical formulae [18] or titles [24] from scientific articles.

## 6 ACKNOWLEDGEMENTS

We are grateful for the support received by Martin Schibel, Andrew Collins and Dominika Tkaczyk in creating the GIANT dataset [19]. We would also like to acknowledge that this research was partly conducted with the financial support of the ADAPT SFI Research Centre at Trinity College Dublin. The ADAPT SFI Centre for Digital Media Technology is funded by Science Foundation Ireland through the SFI Research Centres Programme and is co-funded under the European Regional Development Fund (ERDF) through Grant #13/RC/2106.